\documentclass[letterpaper, 10 pt, conference]{ieeeconf}  % Comment this line out if you need a4paper

\let\labelindent\relax
\usepackage{cite}
\usepackage{amsmath,amssymb,amsfonts}
\usepackage{algorithm}
\usepackage{hyperref}
\usepackage{algpseudocode}
\usepackage{graphicx}
\usepackage{textcomp}
\usepackage{booktabs, multirow}
\usepackage{authblk}
\usepackage{array}
\usepackage{pifont}
\usepackage{xcolor,colortbl}[table]
\usepackage[percent]{overpic}
\usepackage{subfig}
\usepackage{comment}
\usepackage{enumitem}
\usepackage{dsfont}
\usepackage{amsmath}
\usepackage{amssymb}

\usepackage{tablefootnote}
\usepackage{multicol}

\newcommand\blfootnote[1]{%
  \begingroup
  \renewcommand\thefootnote{}\footnote{#1}%
  \addtocounter{footnote}{-1}%
  \endgroup
}

\newcolumntype{F}[1]{>{\raggedright\arraybackslash\hspace{0pt}}p{#1}}%
\newcolumntype{T}[1]{>{\centering\arraybackslash\hspace{0pt}}p{#1}}%
\newcolumntype{P}[1]{>{\centering\arraybackslash\hspace{0pt}\vspace{0pt}}p{#1}}

\def\BibTeX{{\rm B\kern-.05em{\sc i\kern-.025em b}\kern-.08em
    T\kern-.1667em\lower.7ex\hbox{E}\kern-.125emX}}
    
\makeatletter
\renewcommand\AB@affilsepx{, \protect\Affilfont}
\makeatother

\title{\LARGE \bf GSta: Efficient Training Scheme with Siestaed Gaussians for Monocular 3D Scene Reconstruction}

\author[]{Anil Armagan}
\author[]{Albert Sa\`a-Garriga}
\author[]{Bruno Manganelli}
\author[2]{Kyuwon Kim}
\author[]{M. Kerim Yucel}

\affil[]{Samsung R\&D Institute UK (SRUK)}
\affil[2]{Samsung Electronics}

\begin{document}
\maketitle
\thispagestyle{empty}
\pagestyle{empty}

%%%%%%%%%%%%%%%%%%%%%%%%%%%%%%%%%%%%%%%%%%%%%%%%%%%%%%%%%%%%%%%%%%%%%%%%%%%%%%%%
\begin{abstract}
Gaussian Splatting (GS) is a popular approach for 3D reconstruction, mostly due to its ability to converge reasonably fast, faithfully represent the scene and render (novel) views in a fast fashion. However, it suffers from large storage and memory requirements,  and its training speed still lags behind the hash-grid based radiance field approaches (e.g. Instant-NGP), which makes it especially difficult to deploy them in robotics scenarios, where 3D reconstruction is crucial for accurate operation. In this paper, we propose GSta that dynamically identifies Gaussians that have converged well during training, based on their positional and color gradient norms. By forcing such Gaussians into a \textit{siesta} and stopping their updates (\textbf{freezing}) during training, we improve training speed with competitive accuracy compared to state of the art. We also propose an early stopping mechanism based on the PSNR values computed on a subset of training images. Combined with other improvements, such as integrating a learning rate scheduler, GSta achieves an improved Pareto front in convergence speed, memory and storage requirements, while preserving quality. We also show that GSta can improve other methods and complement orthogonal approaches in efficiency improvement; once combined with Trick-GS, GSta achieves up to 5$\times$ faster training, 16$\times$ smaller disk size compared to vanilla GS, while having comparable accuracy and consuming only half the peak memory. More visualisations are available at \url{https://anilarmagan.github.io/SRUK-GSta}.

\end{abstract}

\vspace{-0.75em}
\section{Introduction}
\label{introduction}
3D reconstruction is an integral problem in computer vision, where the aim is to leverage one or multiple images of a scene/object, and lift it to 3D while representing geometry and textures faithfully. With applications in robotics \cite{adamkiewicz2022vision}, edge devices \cite{georgiadis2025cheapnvs}, multimedia \cite{skartados2024finding} and virtual reality \cite{li2023instant}, its industrial importance has also increased over the years. Despite the recent advances in the field, 3D reconstruction is inherently an ill-posed problem, as accurate disentanglement of geometry and texture from 2D images is required, as well as an accurate back-projection of images onto a 3D space is necessary, which is inherently ambiguous. 

\begin{figure}[b!]
\vspace{-4mm}
\centering
\captionsetup[subfigure]{labelformat=empty}
\includegraphics[width=1.0\linewidth]{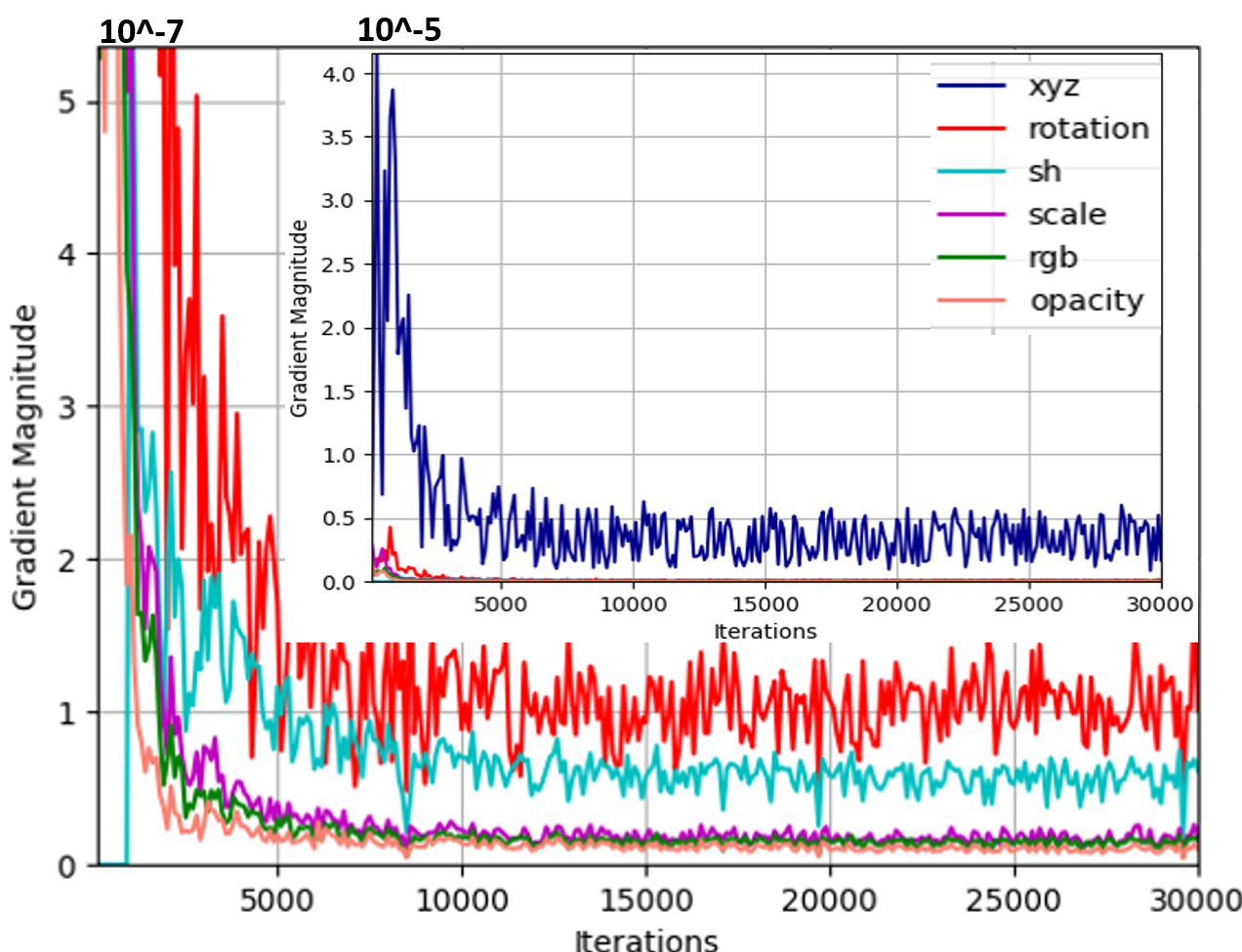}
\vspace{-1.8em}
\caption{\small Mean gradient magnitude of Gaussians during 30000 iterations of training. Training takes $6\times$ longer since hard-coded number of training iterations in 3DGS~\cite{kerbl3Dgaussians}, while a high number of Gaussians are converged in early iterations. Please note the scale difference between the positional and rest of the parameters.}
\label{fig:gaussian_gradients_plt}
\end{figure} 
\blfootnote{This work has been submitted to the IEEE for possible publication. Copyright may be transferred without notice, after which this version may no longer be accessible.}

Following the earlier photogrammetry \cite{cyganek2011introduction} approaches, neural rendering methods have become popular for 3D reconstruction, particularly due to their success in novel view synthesis. Neural Radiance Fields (NeRF), where a scene is encoded into a small neural network, have shown promising results \cite{SyntheticNeRF}, and its variants targeting faster convergence and rendering \cite{muller2022instant,reiser2021kilonerf,yu2021plenoctrees} have emerged as well. Most recently, Gaussian Splatting (GS) \cite{kerbl3Dgaussians} has been proposed, where a scene is represented with many Gaussians, whose parameters are learned with gradient-based optimization. GS based methods are arguably leads the 3D representation/reconstruction technology at the moment, due to their fast rendering speed and reasonable convergence time.

\begin{figure*}[!ht]
\centering
\captionsetup[subfigure]{labelformat=empty}
\includegraphics[width=1.\textwidth]{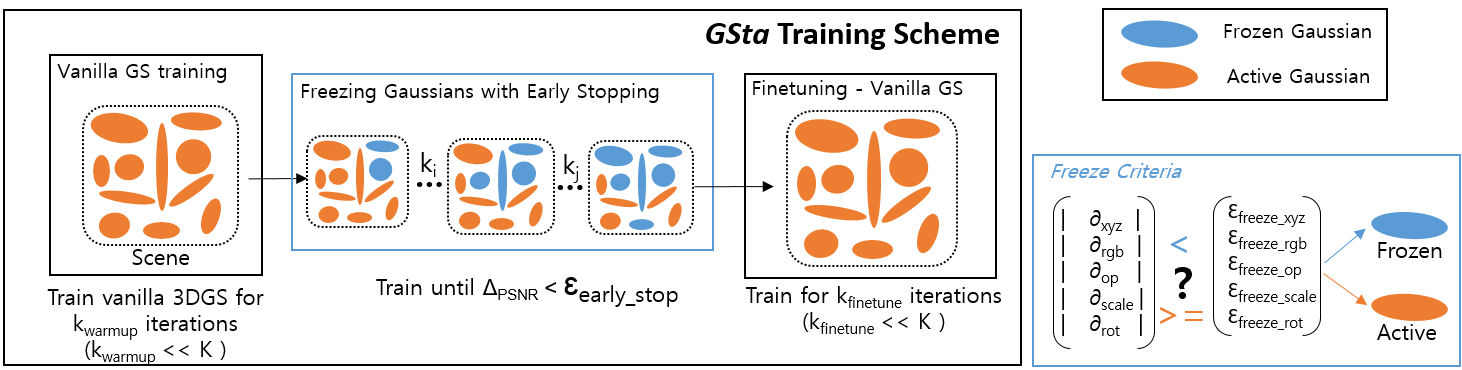}
\vspace{-1.8em}
\caption{\small Our proposed GSta training strategy. During training, we observe the gradients of Gaussian parameters (position \textit{xyz} and color \textit{rgb} used in practice) to decide which Gaussians have converged. We then progressively freeze (e.g. stop training) converged Gaussians, until we either hit our proposed training-set based early stopping criteria, or finish training. We then \textit{unfreeze} all Gaussians and finetune them for a few iterations for global alignment. GSta leads to reduced training time, disk size and peak memory consumption.}
\vspace{-1.2em}
\label{fig:training_scheme}
\vspace{-2mm}
\end{figure*} 

Despite its success, GS suffers from several shortcomings. First, since it represents the scenes with many Gaussians, ranging up to several millions, it has large storage requirements. Second, these many Gaussians lead to high maximum memory consumption during training, limiting the feasibility of training on devices with resource constraints. In the same vein, optimization of these many Gaussians lead to a somewhat acceptable, but definitely improvable training time. Several methods have been proposed targeting these shortcomings, such as ones reducing the number of Gaussians \cite{fan2024lightGaussian,hanson2024pup3dgs,morgenstern2024compact,papantonakis2024reducing,ali2024trimmingfat}, number of attributes \cite{papantonakis2024reducing, lee2024compact, fan2024lightGaussian}, and directly predicting Gaussians by leveraging diffusion priors \cite{chen2024text}. However, there is still room for improvement on multiple axes regarding efficiency as well as fidelity.

Our work aims to advance the Pareto front of efficiency in Gaussian Splatting methods, aiming to improve training time, storage and memory requirements, while keeping the accuracy competitive.  First, based on the observation that each Gaussian has different convergence traits during training (e.g. some converging earlier than others), we exploit this to improve training efficiency. An insignificant number of Gaussians remain with high gradients after early stages of training. We infer this by visualising the mean gradient magnitude, Figure~\ref{fig:gaussian_gradients_plt}, and the maximum gradient magnitudes over all Gaussians. While the mean gradient magnitudes converge around 5K iterations, we could measure the variance of the magnitudes varies in later stages of the training. Based on the  norms of the color and positional gradients of Gaussians during training, we force some of the Gaussians (which have gradient norms below a threshold) to take a \textit{siesta}, and freeze them for the next iterations of the training. This gradually decreases the overall gradient flow to calculate and optimize, improving training efficiency. Second, we introduce a simple but effective early stopping method for GS training. Unlike methods which use testing images for early stopping validation, we observe that using a subset of training images work equally well, if not slightly better. This facilitates a \textit{fairer} early stopping as we do not tune model weights on the test images. Third, we introduce improvements for a more stable training, such as learning rate schedulers and using dynamic thresholds for gradient thresholding within adaptive densification and freezing. We also investigate the use of gradients from diffused colors together with the positional gradients which improve the overall effectiveness of the training wit a lower number of Gaussians.

We evaluate GSta on multiple datasets and integrate it with other methods, showing that it can improve the vanilla model as well as another efficient GS method (Trick-GS~\cite{armagan2025trickgs}). Our approach works well with orthogonal techniques for improving efficiency and our contributions are as follows:
\begin{itemize}
\vspace{1mm}
    \item We propose GSta, a gradient-based method that dynamically identifies Gaussians that are close to convergence, and freezes them for the rest of the training. This helps improve training speed, without hurting accuracy.
    \item Alongside GSta, we introduce a global finetuning stage, where we force all Gaussians out of their \textit{siesta} to globally align them for better results.
    \item We propose a training-set based early stopping method, where we end the training if performance on a randomly chosen subset of training images saturate. This helps improve the training speed.
    %, and also is a fairer way of validating as it does not use test images.
    \item We show that GSta works well with existing GS methods, as well as other efficiency-improving techniques. Our results on three datasets show substantial gains in training time, disk size and peak memory consumption, while having competitive accuracy.
\end{itemize}

\section{Related Work} \label{related_work}
\noindent \textbf{Neural 3D Reconstruction.}  NeRF approaches, where the appearance and geometry of a scene are encoded into the weights of a small neural network, have been popularized thanks to their compact representation as well as high-quality results. Despite the advances made in the efficiency of NeRF methods \cite{muller2022instant,reiser2021kilonerf,yu2021plenoctrees}, NeRFs rely on the expensive ray marching process, which inevitably increases the runtime. Gaussian Splatting methods \cite{kerbl3Dgaussians} does away with ray marching; they represent the scene with 3D Gaussian primitives and perform blending in screen space, speeding up the rendering process quite significantly. Despite their popularity, GS methods suffer from high storage and training memory requirements, since scenes are represented with over million of Gaussian primitives. This also translates to sub-optimal training times, and rendering speeds that can still be improved.

\noindent \textbf{Improving Storage Efficiency.} The storage footprint of GS methods rely on two major factors; number of Gaussians used to represent the scene and the 59 primitives that make up a Gaussian. There are several methods targeting to address the former by assigning a significance score \cite{liu2024maskgaussian,hanson2024pup,zoomers2024progs} to each Gaussian to eliminate the insignificant ones; \cite{fan2024lightGaussian} computes significance on how frequently a Gaussian comes up in a training view, \cite{lee2024compact} learns a significance value during training, \cite{seo2024flod} removes Gaussians with significant overlap to its neighbors, \cite{papantonakis2024reducing }  removes 3\% of lowest-opacity Gaussians, \cite{ali2024trimmingfat} performs a fine-tuning stage for pruning and \cite{zhang2024gaussianspa} explicitly enforces sparsity among Gaussians. Some methods address the latter, where \cite{papantonakis2024reducing, lee2024compact, fan2024lightGaussian,shin2025locality,wang2024sg} propose to reduce/mask/replace spherical harmonic bands and \cite{lee2024compact} leverages a hash-grid to compress color parameters. Parallel to these two lines of work, attribute compression via Vector Quantization \cite{lee2024compact, fan2024lightGaussian, niedermayr2024compressed}, bit-quantization \cite{wang2024end, niedermayr2024compressed} and leveraging entropy constraints \cite{girish2024eagles, niedermayr2024compressed,ali2024elmgs,xie2024mesongs,liu2024hemgs} are shown to be effective for storage footprint reduction. A final line of work focus on structural relationships between Gaussians in a scene, and propose anchors \cite{lu2024scaffold, chen2024hac, wang2024contextgs, liu2024compgs} from which Gaussians can be derived efficiently. Mapping Gaussian attributes to 2D grids \cite{morgenstern2024compact}, adopting efficient data structures like octrees \cite{ren2024octree} and tri-planes \cite{wu2024implicit}, using Gaussians' spatial positions for improved organization \cite{fang2024minisplatting,shin2025locality} and leveraging selective rendering \cite{seo2024flod} are some examples.

\noindent \textbf{Improving Training Efficiency.} Similar to NeRFs, Gaussian Splatting methods often require per-scene training, which makes training efficiency a critical factor. Most methods that improve storage efficiency also tend to improve training efficiency, therefore we will not mention them separately here. Other methods, such as \cite{durvasula2023distwar}  accelerating atomic operations in raster-based differential rendering, separation of SH bands into different tensors to load to rasterizer separately \cite{mallick2024taming3dgs}, densification based on position and appearance criteria \cite{lu2024turbogs}, organization of Gaussians into manageable groups \cite{wang2024faster} and introduction of visibility culling \cite{fang2024mini2} have shown training speed improvements. 

A similar line of work to ours is \cite{lu2024turbogs}, where authors leverage positional or appearance gradients for densification. Our work differs from theirs in several key aspects; we i) use positional and appearance gradients jointly, ii) use them for freezing Gaussians during training, not just for densification, iii) propose a finetuning stage for global optimization of frozen Gaussians, iv) leverage PSNR values of a subset of training images for early stopping and v) propose other minor tricks, such as using new optimizers and learning rate schedulers. GSta is a plug-and-play improvement that can improve several GS methods and it complements other efficiency-improving approaches.

\section{Methodology}
\label{methodology}
In this section, we first give preliminary information on 3DGS and later describe our method in detail.

\subsection{Preliminaries}
3DGS performs 3D reconstruction by learning a number of 3D Gaussian primitives, which are rendered in a differentiable volume splatting rasterizer \cite{zwicker2001ewa}. A scene/object is often represented with up to several millions of Gaussians, where each Gaussian consists of 59 parameters. The 2D Gaussian $G_{i}'$ projected from a 3D Gaussian $G_i$ is defined as
%\vspace{-0.6em}
\begin{equation}
\label{eq:2d-Gaussian-proj}
   G_{i}'(x) = e^{-\frac{1}{2}(x-\mu_{i}')^T{\Sigma_{i}'}^{-1}(x-\mu_{i}')},\vspace{-0.6em}
\end{equation}
\noindent
where \(\mu_{i}\) is the 3D position, \(x\) is the position vector,  and \(\Sigma_{i}\) is the 3D covariance matrix which is parameterized via a scaling matrix \(S\) and a rotation matrix (represented in quaternions \(q\)). $'$ indicates the 2D re-projection of respective parameters. Each Gaussian has an opacity (\(\alpha \in [0, 1]\)), a diffused color and a set of spherical harmonics (SH) coefficients (up to 3 bands) that represent view-dependent color information. Color \(C\) of each pixel is determined by \(N\) Gaussians contributing to that pixel, where their colors are blended in a sorted order within the volumetric rendering regime. Color C is calculated as 
% \vspace{-0.6em}
\begin{equation}
\label{colorblending}
\small
    C = \sum_{i \in N} c_i \alpha_i  \prod_{j=1}^{i-1} (1 - \alpha_j),\vspace{-0.6em}
\end{equation}
\noindent
where \(c_i\) and \(\alpha_i\) are the view-dependent color and opacity of a Gaussian. respectively. The standard L1 reconstruction loss and SSIM loss is used for training, and Gaussians are pruned and densified based on their gradient norms and opacity values.

Gaussians are initialized from a point-cloud calculated from \(SfM\) \cite{schoenberger2016sfm} methods, and they are iteratively updated based on the gradient descent algorithm, where training views of a scene are rendered. The attributes of Gaussians are updated with the calculated gradients based on $\ell_1$ norm of the rendered output, $\mathcal{L}_1$, and its structural similarity, $\mathcal{L}_{\text {D-SSIM }}$ with the ground-truth image. The final loss is calculated as in Eq.\ref{loss3dgs} to obtain a well represented scene reconstruction until $K^{th}$ training iteration. The overall loss is defined as 
\begin{equation}
\label{loss3dgs}
\mathcal{L}=(1-\lambda_{SSIM}) \mathcal{L}_1+\lambda_{SSIM} \mathcal{L}_{\text {D-SSIM }},
\end{equation}
where $\lambda_{SSIM}=0.2$.

\subsection{Efficient training scheme}
\label{sec:training_scheme}
Our proposed scheme is based on the observation that Gaussians have different convergence rates during training. Figure~\ref{fig:gaussian_gradients_plt} shows the mean magnitude of all Gaussians during the training of "bicycle" scene of MipNeRF-360~\cite{MipNeRF360} dataset. It can be observed that most Gaussians do converge early, however the densification stage still continues to add more Gaussians, due to some Gaussians (e.g. ones that did not yet converge) still having high gradients. This supports the fact that a large number of Gaussians do not need to be trained (e.g. frozen) further, and we can just continue training (e.g. unfrozen) the rest that have high gradients. The frozen Gaussians should ideally be unfrozen at some point, since the reconstruction quality will change with the updated Gaussians, and the frozen Gaussians might need to be updated to align with the new fitting. We detail our  overall algorithm in Algorithm \ref{alg:algorithm}. 

\subsubsection{Leveraging gradients for efficient GS training}
\label{sec:freeze}
Vanilla 3DGS uses positional parameters and their gradients for its densification process. In GSta, we use a similar strategy for efficient training, and take action depending on magnitude of the chosen parameters' gradients. More specifically, GSta freezes the parameters of Gaussians' during training, if the magnitude of the parameters' gradient is lower than a threshold. We use positional and diffused color parameters' gradients for this purpose. Additionally, we freeze Gaussians only when they have low magnitude gradients on both color and positional parameters, and not just one of them.

Vanilla GS uses fixed threshold values to decide which Gaussians need to be densified or not, based on their gradient magnitudes. GSta, on the other hand, dynamically changes the thresholds that are used to decide which Gaussians to be frozen. We enforce the thresholds for color and positional parameters $p \in \{xyz, rgb\}$ to be between $\lambda_{freeze,1} * \epsilon^{p}_{k,freeze}$ and $\lambda_{freeze,2} * \epsilon^{p}_{k,freeze}$, where $\lambda_{freeze,1} =0.5, \lambda_{freeze,2} = 1.5$ . We gradually increase the thresholds proportional to current training iteration $k$ until the training hits an early stopping criteria (see Sec.~\ref{sec:early_stop_lr}) or until the adaptive densification stops, as we aim to freeze less Gaussians in early stages of the training and freeze more as the number of Gaussians increase.

A Gaussian's gradients are calculated during backpropagation and corresponding weights are updated only if Gaussian's freeze criteria, $freeze^{p}_{i}$ is set to false, which is decided as
\begin{equation}
  freeze^{p}_{i} =\left\{\begin{matrix}
    True, & \|grad^{p}_{i}\| < \epsilon^{p}_{k,freeze} \\
    False, & \textup{otherwise}, \\
\end{matrix}\right.
\label{eq:freeze_criteria}
\end{equation}
where $k$ is the current training iteration and $\|grad^{p}_{i}\|$ is magnitude of the gradient for $p^{th}$ parameter of $i^{th}$ Gaussian. $\epsilon^{p}_{k,freeze}$ is a threshold chosen for the current training iteration $k$ and for the specific Gaussian parameter $p$, which is calculated as
\begin{equation}
\begin{aligned}
\epsilon^{p}_{k,freeze} = \lambda_{freeze,1} * \epsilon^{p}_{k-1,freeze} + \\ k/K * \lambda_{freeze,2}* \epsilon^{p}_{k-1,freeze}
\label{eq:freeze_threshold}
\end{aligned}
\end{equation}

During training, GSta freezes a Gaussian, $G_{i}$ only if both $freeze^{xyz}_{i}$ and $freeze^{rgb}_{i}$ is $True$.
\begin{equation}
freeze_{i} =\left\{\begin{matrix}
    True, & freeze^{xyz}_{i} \ \text{and} \ freeze^{rgb}_{i} \\
    False, & \textup{otherwise}, \\
\end{matrix}\right.
\label{eq:freeze_gaussian_criteria}
\end{equation}

where $freeze_{i}$ is the freeze map at iteration $i$ (see Figure \ref{fig:training_scheme} for a visualization).

\subsubsection{Early stopping}
\label{sec:early_stop_lr}
Most GS models are trained with fixed number of training iterations. An efficient GS training strategy requires to be aware of convergence. Therefore, we adopt early stopping strategy based on accuracy metrics, such as $PSNR$. We enforce the early stopping of GSta's freezing strategy and enable all Gaussians to be updated for $K_{f}$ number of finetuning iterations. This early stopping is invoked if the change in $PSNR$ metric for the $k^{th}$ training iteration, $\Delta PSNR_{k}$, is lower than a threshold, $\epsilon_{PSNR}$ value.
\newcommand\lreqn[2]{\noindent\makebox[\textwidth/3]{$\displaystyle#1$\hfill#2}\vspace{1ex}}

However, calculating the $PSNR$ at every training iteration is costly and therefore, we update the metric every $K_{PSNR}$ iterations. We also use a patience factor (wait count) of 1 and require the early stopping criteria to be met twice to define a more reliable convergence criteria. We define the early stopping criteria as
\begin{equation}
\mathrm{early\_stop} = 
\begin{cases}
    \text{True,} & \Delta PSNR_{k} < \epsilon_{PSNR} \ \text{and} \\
                 & \Delta PSNR_{k - K_{PSNR}} < \epsilon_{PSNR}, \\
    \text{False,} & \text{otherwise},
\end{cases}
% \right.
\label{eq:early_stop_criteria}
\end{equation}

Note that we do not use the test images for early stopping, which means we do not tune the model based on the test accuracy. Instead, we observe that using the training images for this purpose gives equally good results, which makes it \textit{fairer} than tuning model parameters on the test set itself. 
\subsubsection{Learning Rate Scheduling}
\label{sec:lr_scheduling}
Vanilla GS is trained with a learning rate for the positional parameters with a cosine decay factor and fixed learning rates for the other parameters. We observe that the final accuracy is highly sensitive to the learning rates and their decay factor. Therefore, we use a learning rate scheduler, which helps us guarantee convergence with the early stopping criteria. More specifically, we use a plateau learning rate scheduler based on the $PSNR$ metric and decay the learning rate of the positional Gaussian parameter with a factor, $d$. The plateau is decided based on threshold value of $\epsilon_{LR-PSNR}$ and a patience factor of 1. 

\subsubsection{Rasterizer \& optimizer}
The vanilla rasterizer in 3DGS is implemented to calculate gradients propagated from the pixels to the Gaussians. However, we need to propagate the gradients per Gaussian during the backward pass, since GSta adopts per splat based freezing strategy. Taming-3DGS~\cite{mallick2024taming3dgs} proposes a parallelization scheme per tile over the 2D splats. Their rasterizer makes better use of storing per-splat state and continually exchange per-pixel states between the threads. Hence, we make use of the per splat based rasterizer~\cite{mallick2024taming3dgs} and modify it to use our freeze maps. The modified rasterizer bypasses the gradient calculation in the backward pass, if the Gaussian's, $G_{i}$, freeze map label, $freeze_{i}$, is set to $False$. 

Since the gradients are not calculated for some Gaussians, the optimizer is not going to update any parameters of those Gaussians. Therefore, we can also make use of a modified optimizer~\cite{mallick2024taming3dgs} to only update the Gaussian parameters if $freeze_{i}$ is set to $True$. Note that both these rasterizer and optimizer saves computational resources, which improves training efficiency and speed.

\subsection{Adding Bag of Tricks for Increased Efficiency}
Our proposed algorithm can reduce the training time by $6\times$, which includes the Gaussian freezing with early stopping strategy, learning rate scheduling and by using the proposed rasterizer and optimizer. We also make use of other, ideally orthogonal efficiency improvement methods along with GSta to push the frontier even further. We mostly use the tricks of Trick-GS \cite{armagan2025trickgs}; readers are referred to the original work for further details. We explain each trick briefly below.

\noindent\textbf{Pruning with Volume Masking.}
\cite{morgenstern2024compact} learns masks for pruning the Gaussians with low scale and opacity. Hard masks, $M\in \{0,1\}^{N}$, are learned for $N$ Gaussians and applied on their opacity and non-negative scale attributes by introducing a soft mask parameter, $m\in \mathbb{R}^{N}$ for each Gaussian. Introduced mask parameters are removed at the end of the training and do not add an extra storage cost. Hard masks can be extracted with $M_n = \operatorname*{sg}(\mathds(\sigma(m_n) > \epsilon) - \sigma(m_n)) + \sigma(m_n)$, where $sg$ is the stop gradient operator and $\sigma$ is the sigmoid function. Soft masks are learned as a logistic regression problem.

\noindent\textbf{Pruning with Significance of Gaussians.}
We adopt the significance based Gaussian pruning strategy proposed in~\cite{fan2024lightGaussian}. Significance of the Gaussians are determined by considering all the rays passing through each training pixel and then by calculating how many times each Gaussian is hit. Scale and opacity of each Gaussian  also contributes to the final score for each Gaussian. Finally, the scores are used to prune percentile of the whole set, starting with the Gaussians with lowest significance. This pruning strategy is applied only twice during training runs, since the score calculation for the whole set is costly.

\noindent\textbf{Spherical Harmonic (SH) Masking.}
Almost 76\% of Gaussian parameters are SH parameters of 3 SH bands. Therefore, they represent the arguably biggest bottleneck for achieving low storage requirements. We choose a similar strategy to Gaussian masking~\cite{morgenstern2024compact} and SH bands pruning~\cite{wang2024end} for each Gaussian. Instead of learning a single soft mask parameter for each Gaussian, we learn 3 parameters corresponding to SH bands of a Gaussian. Instead of pruning SH bands, they are set to zero and used within the rasterizer. At the end of training, SH bands that are masked out are pruned and not saved, improving storage footprint. Note that we use SH masking only with $GSta+TrickGS+small$ (see Sec \ref{sec:sota_comparison}). 

\noindent\textbf{Progressive Training.}
Progressive training strategies help lower storage and training time by starting from a lower resolution image representation. Gradually increasing the resolution~\cite{yang2024spec, girish2024eagles} lets the optimization focus on low frequency details first and high frequency details at later iterations. We adopt a similar trick and starting from a lower resolution, we increase the resolution size with a cosine scheduler until 6K iterations. 

\noindent \textbf{Blur.} Another way to make the optimization focus on low frequency details first is adding Gaussian blur~\cite{girish2024eagles} on the image. We add Gaussian blur on the original image by starting with a larger kernel size and gradually decay it with a cosine scheduler and remove the noise until 6K iterations.

\noindent\textbf{Accelerated SSIM loss calculation.}
Another trick for training time efficiency is to modify $SSIM$ loss calculation with optimized CUDA kernels~\cite{mallick2024taming3dgs}, where each 2D convolution kernels with in the calculation is replaced with two 1D kernels. Later a fused kernel from the output of 1D convolutions are used to calculate $SSIM$.

\section{Experimental Results}
\label{experiments}

% \vspace{-0.5em}
We first include details of our experimental setup and implementation, and then discuss our experimental evaluation and comparison with state of the art methods. 

\subsection{Experimental Setup and Implementation Details}

\noindent \textbf{Metrics and datasets.} We evaluate our method on various bounded and unbounded indoor/outdoor scenes; Mip-NeRF-360 \cite{MipNeRF360}, Tanks\&Temples \cite{TanksAndTemples} and DeepBlending \cite{DeepBlending} are used in our experiments. Following \cite{kerbl3Dgaussians}, we use the pre-computed point clouds and camera poses for training and use every $8^{th}$ image for evaluation. We use the commonly used PSNR, SSIM and LPIPS metrics \cite{zhang2018unreasonable} for evaluating GSta. We train and evaluate our method on a single NVIDIA RTX 3090 and use \textit{torch.cuda.Event} function to report training time (including densification), and provide FPS values as the average of 50 runs using \textit{torch.utils.benchmark}. Note that we measure training times and FPS values on the same hardware, including Mini-Splatting-v2 and Taming-3DGS~\cite{fang2024mini2, mallick2024taming3dgs}.  

\begin{table*}[!t]
\centering
\resizebox{1.0\linewidth}{!}{
\begin{tabular}{l|cccT{0.07\textwidth}T{0.05\textwidth}T{0.08\textwidth}T{0.06\textwidth}|cccT{0.07\textwidth}T{0.05\textwidth}T{0.08\textwidth}T{0.06\textwidth}}
\hline
\rowcolor{lightgray} Dataset      & \multicolumn{7}{c}{Mip-NeRF 360}                & \multicolumn{7}{c}{Tanks\&Temples}              \\
\hline
% \cmidrule(lr){2-9}\cmidrule(lr){10-17}
 Method & PSNR ↑ & SSIM ↑ & LPIPS ↓ & Storage ↓ (MB) & Time ↓ (min) &
 Max-\#GS ↓ $\times1000$ & \#GS ↓ $\times1000$ & PSNR ↑ & SSIM ↑ & LPIPS ↓ & Storage ↓ (MB) & Time↓ (min) &
 Max-\#GS ↓ $\times1000$ & \#GS ↓
$\times1000$ \\ 
% \midrule

\hline

\rowcolor{lightgray} \textsuperscript{*}INGP-base \cite{muller2022instant} & 25.30	& 0.671	& 0.371	& 13 & 5.62	 & - & - & 21.72 & 0.723 & 0.330 & 13 & 5.43  & - & -\tabularnewline

\textsuperscript{*}INGP-big \cite{muller2022instant} & 25.59	& 0.699	& 0.331	& 48 & 7.50	& -  & - & 21.92 & 0.745 & 0.305 & 48 & 6.98  & - & -\tabularnewline

\rowcolor{lightgray} \textsuperscript{*}Turbo-GS~\cite{lu2024turbogs} & 27.38 & 0.812 & 0.210 & 240 & 5.47  & - & 490 & 23.49 & 0.841 & 0.176 & - & 4.35 & - & -\tabularnewline

\hline\hline

3DGS~\cite{kerbl3Dgaussians} & 27.56 & 0.818 & 0.202 & 770 & 23.83  & 3265 & 3255 & 23.67 & 0.845 & 0.178 & 431 & 14.69  & 1824 & 1824\tabularnewline

% Mini-Splatting~\cite{fang2024minisplatting} & 27.25 & 0.820 & 0.219 & 118 & 21.92 & 364 & 4235 & 496 & 23.23 & 0.833 & 0.202 & 48 & 12.34 & 452 & 4291 & 203\tabularnewline

\rowcolor{lightgray} Mini-Splatting-v2~\cite{fang2024mini2} & 27.37 & 0.821 & 0.215 & 139 & 3.93  & 4744 & 618 & 23.16 & 0.842 & 0.185 & 84 & 2.35  & 3890 & 358\tabularnewline

Taming-3DGS~\cite{mallick2024taming3dgs} & 27.25 & 0.795 & 0.260 & 151 & 6.46  & 670 & 670 & 23.73 & 0.836 & 0.210 & 72 & 4.02  & 319 & 319 \tabularnewline

\rowcolor{lightgray} Trick-GS~\cite{armagan2025trickgs} & 27.16 & 0.802 & 0.245 & 39 & 15.41   & 1369 & 830 & 23.48 & 0.830 & 0.209 & 20 & 10.56  & 696 & 443\tabularnewline
\hline\hline
GSta & 27.21 & 0.812 & 0.227 & 304 & 7.89  & 2740 & 2699 & 22.90 & 0.833 & 0.203 & 165 & 3.99   & 1511 & 1468\tabularnewline
% 23.19 & 0.837 & 0.197 & 166 & 5.21 & -  & 1891 & 1891\tabularnewline

\rowcolor{lightgray}GSta+Trick-GS+small & 27.13 & 0.807 & 0.242 & 48 & 5.30  & 1774 & 1148 & 23.02 & 0.830 & 0.207 & 28 & 4.37   & 920 & 727\tabularnewline
% 22.87 & 0.824 & 0.220 & 27 & 3.26 & 242  & 905 & 702\tabularnewline

GSta+Trick-GS & 27.14 & 0.807 & 0.241 & 130 & 4.84  & 1781 & 1159 & 23.13 & 0.832	& 0.202	& 103	& 2.89 & 1252 & 914
\vspace{-1.5em}
\\\hline
\end{tabular}}
\vspace{-0.3em}
\caption{\small Quantitative evaluation on MipNeRF 360 and Tanks\&Temples datasets. Results marked with ``$^*$" are taken from the corresponding papers. Results between the double horizontal lines are from retraining the models on our system. GSta can reconstruct scenes with low training time and very low disk space requirements, while achieving competitive accuracy.}
\vspace{-1.5em}
\label{tab:qual1_2}
\end{table*}

\begin{table}[!t]
\centering
% \vspace{-0.4em}
\resizebox{1.\linewidth}{!}{
\begin{tabular}{l|cccT{0.07\textwidth}T{0.05\textwidth}T{0.08\textwidth}T{0.06\textwidth}}
\hline
\rowcolor{lightgray} Dataset      & \multicolumn{7}{c}{Deep Blending}\\ %\cmidrule(lr){2-9}
\hline
Method & PSNR ↑ & SSIM ↑ & LPIPS ↓ & Storage ↓ (MB) & Time ↓ (min) & Max-\#GS ↓ $\times1000$ & \#GS ↓ $\times1000$ \\
% \midrule
\hline

\rowcolor{lightgray} \textsuperscript{*}INGP-base~\cite{muller2022instant} & 23.62 & 0.797 & 0.423 & 13 & 6.52 & - & -\tabularnewline

\textsuperscript{*}INGP-big~\cite{muller2022instant} & 24.96 & 0.817 & 0.390 & 48 & 8.00  & - & -\tabularnewline

\rowcolor{lightgray} \textsuperscript{*}Turbo-GS~\cite{lu2024turbogs} & 30.41 & 0.910 & 0.240 & - & 3.24 &  - & -\tabularnewline
\hline\hline

3DGS~\cite{kerbl3Dgaussians} & 29.46 & 0.899 & 0.266 & 664 & 25.29  & 2808 & 2808 \tabularnewline

% \rowcolor{lightgray} Mini-Splatting~\cite{fang2024minisplatting} & 29.78 & 0.906 & 0.251 & 80 & 16.53 & 367 & 4521 & 339\tabularnewline

\rowcolor{lightgray}Mini-Splatting-v2~\cite{fang2024mini2} & 30.19 & 0.912 & 0.240 & 73 & 3.16  & 5255 & 651\tabularnewline

Taming-3DGS~\cite{mallick2024taming3dgs} & 30.07 & 0.904 & 0.270 & 33 & 4.52  & 294 & 294\tabularnewline

\rowcolor{lightgray}Trick-GS~\cite{armagan2025trickgs} & 29.46 & 0.899 & 0.260 & 25 & 13.11  & 1308 & 639\tabularnewline
\hline\hline
GSta & 29.40 & 0.906 & 0.251 & 296 & 7.37 &  2629 & 2629\tabularnewline

\rowcolor{lightgray}GSta+Trick-GS+small & 29.33 & 0.905 & 0.260 & 48 & 4.44  & 1877 & 1399\tabularnewline

GSta+Trick-GS & 29.41 & 0.906 & 0.257 & 158 & 4.68  & 1887 & 1405
\\\hline
\end{tabular}}
% \vspace{-0.3em}
\caption{\small Quantitative evaluation on Deep Blending dataset. Results marked with $^*$ are taken from the corresponding papers. Results below the double horizontal line are trained and evaluated on our system. }
\vspace{-1.3em}

\label{tab:qual_db2_v2}
\vspace{-1.5em}
\end{table}

\algnewcommand{\LineComment}[1]{\Statex \hskip\ALG \(\triangleright\) #1}

\begin{algorithm}
\caption{GSta: Efficient Training Iteration for 3DGS.}
\footnotesize

\begin{itemize}[leftmargin=*]

\item \textcolor{gray}{G: Set of all Gaussian parameters, gt\_image: ground-truth image,  cam\_pose: camera pose of gt\_image.}
\item \textcolor{gray}{eval\_image: evaluation images, eval\_cam\_pose: camera poses of eval\_image, eval\_freq: frequency of evaluation.}
\item \textcolor{gray}{total\_iter: total train iterations, iter: current train iteration, psnrs: list of previous PSNRs.}
\item \textcolor{gray}{stop\_p: early stopping patience, stop\_eps: early stopping threshold, max\_stop\_it: early stopping max number of iterations.}
\item \textcolor{gray}{eps\_f\_xyz, eps\_f\_rgb: initial thresholds for xyz and rgb gradients for freezing.} 
\item \textcolor{gray}{finetune: current finetuning state, ft\_iter\_rem: remaining finetuning iterations.} 

\end{itemize}

\begin{algorithmic}[1]

\While{$iter < total\_iter$ or $ft\_iter\_rem != 0$}

\Comment{\textcolor{gray}{Train until max training or finetuning iteration is reached.}}

\State {$gt\_image, cam\_pose \gets sample\_batch()$; \Comment{\textcolor{gray}{Sample batch.}}}

\State {$eps\_f\_xyz, eps\_f\_rgb \gets$  \Comment{\textcolor{gray}{Freezing threshold value update.}}
 \\ 
\quad\quad\quad\quad$update\_epsilons(iter, total\_iter, eps\_f\_xyz, eps\_f\_rgb)$;}

\State {$early\_stop \gets$  \Comment{\textcolor{gray}{Early stopping based on PSNR values.}}\\
\quad\quad\quad\quad$is\_early\_stop(psnrs, iter, stop\_p, stop\_eps, max\_stop\_it)$;}

\If {$early\_stop = False$ AND $finetune = False$}

    \State {$freeze\_map \gets$ 
    \Comment{\textcolor{gray}{Update freeze map with new thresholds.}}\\
    \quad\quad\quad\quad$update\_freeze\_map(G, eps\_f\_xyz, eps\_f\_rgb)$;}
    
\ElsIf {$finetune = False$}

\Comment{\textcolor{gray}{Early stop is True. Unfreeze all Gaussians, activate finetuning.}}
    \State {$freeze\_map \gets reset\_freeze\_map(G)$;}
    
    \State {$finetune \gets True$;}
    
\EndIf

\Comment{\textcolor{gray}{Periodically validate model to get PSNRs for early stopping.}}
\If {$iter \% eval\_freq = 0$ and $finetune = False$}
\State $psnrs \gets validate\_model(G, eval\_ims, eval\_cam\_pose)$;
\EndIf
\State $lr \gets LR\_scheduler(psnrs)$;
 \Comment{\textcolor{gray}{Update LR.}}

\State $render\_image \gets rasterize(G, freeze\_map, cam\_pose)$;  

\State $loss \gets loss\_fcn(gt\_image, render\_image)$; \Comment{\textcolor{gray}{Calculate loss.}}

\State $ optimizer\_step(G, lr, loss, freeze\_map)$; 

\State $G \gets adaptive\_densification(G)$;
\Comment{\textcolor{gray}{Densify.}}

\State $G \gets opacity\_reset(G)$;
\Comment{\textcolor{gray}{Reset opacities.}}
\If {$finetune = True$ AND $ft\_iter\_rem > 0$}    
    \State $ft\_iter\_rem \gets ft\_iter\_rem-1$;
    
\ElsIf{$early\_stop \gets False$ OR $iter < total\_iter$}

    \State $iter \gets iter+1$;

\EndIf

\EndWhile

\vspace{-1mm}
\end{algorithmic}
\label{alg:algorithm}
\end{algorithm}

\noindent \textbf{GSta Details.} 
GSta dynamically adjusts the thresholds used for the densification and freezing Gaussians. More specifically, thresholds are proportionally adjusted to the current training iteration and are scaled between $[0.5, 1.5]$ starting with their initial values. Initial thresholds for the densification are set to $eps\_den\_xyz=0.0002$ and $eps\_den\_rgb=0.000001$. Initial thresholds for freezing Gaussians are set to $eps\_f\_xyz=0.00003$ and $eps\_f\_rgb=0.0001$. The thresholds are chosen once by observing the convergence of each parameters' gradient magnitudes and applied the same on all scenes. Freeze map is updated according to the current parameters' thresholds every $250$ iterations and frozen Gaussians are accumulated until our algorithm meets early stopping criteria. It implies that if a Gaussian is frozen in the previous iterations it will stay frozen in later iterations. GSta unfreezes some Gaussians only if they would be pruned within the densification stage. We also reset the freeze map at every $k_{f_{reset}}=2K$ iterations and do not start re-freezing until $k_{f_{wait}}=500$ iterations pass to give chance for some Gaussians to be unfrozen later. 

Our algorithm early stops and starts finetuning the representation if the change in PSNR is less than $0.2dB$ with a patience of $1$ PSNR calculations, and we update models' current PSNR every $1K$ iterations until the freezing complete. GSta trains without freezing for $k_{warmup}=3K$ iterations and activates the freezing algorithm between $[3K, \min(10K, early\_stop\_iter)]$ iterations. Although we train in full fp32 precision, we save and evaluate our models in reduced fp16 precision, which allows us to lower the storage size for practically no loss in accuracy. 

 GSta uses the same learning rates as the vanilla model, except the opacity learning rate which is $0.025$. Plateau learning rate schedulers are implemented with a learning rate decay factor of $0.75$, patience of $1$ and epsilon of $1$.
 
 We densify every 100 iterations starting from iteration $(500,\min(10K, early\_stop\_iter))$, and prune unnecessary Gaussians with the aforementioned pruning strategies. Similarly, opacity reset is applied from iterations $[3K,\min(10K, early\_stop\_iter)]$ with a step size of $3K$. 
 
 We additionally use a trick introduced in $Taming-GS$~\cite{mallick2024taming3dgs} and replace the activation function of opacity from "sigmoid" to "absolute" activation function when finetuning of our method is activated. The regular capped Gaussian primitives are converted  to high-opacity Gaussians which is shown to represent opaque surfaces with less number of Gaussians. Following Turbo-GS~\cite{lu2024turbogs}, we adopt gradients from diffused color parameters together with positional parameters. They are used both for adaptive densification and in our case, also for freezing. Gradients from diffused colors are activated 20\% of the time to be used in densification and they are always used together with the positional based gradients for freezing Gaussians.
 
\noindent \textbf{Trick-GS Details.} We get inspiration from Trick-GS\cite{armagan2025trickgs} for scheduling image resolutions, blurring and significance score based pruning. Starting from an image resolution with scale $0.175$, we gradually increase it back to the original resolution until iteration $6K$ using a cosine scheduler. Similarly, Gaussian kernels are initialized to $5\times5$, and kernel size is adapted to correspond $0.25$ of the image scales. Kernel size is gradually lowered with a $\log$ scheduler to remove the noise until iteration $6K$. Significance based pruning is applied at iterations $4K$ and $7K$ with $75\%$ of removal factor and a decay of $0.8$. SH masks are learned only during finetuning iterations. We learn adaptive GS mask pruning from iteration 1K to 10K and prune Gaussians every 1K iterations. Please refer to~\cite{armagan2025trickgs} for the rest of the parameters as we keep the rest of settings the same.

\vspace{-2mm}
\subsection{Performance Evaluation}
\vspace{-1mm}
\label{sec:sota_comparison}
We compare GSta with four state of the art approaches, Mini-Splatting-v2~\cite{fang2024mini2}, Taming-GS~\cite{mallick2024taming3dgs}, Trick-GS~\cite{armagan2025trickgs} and finally Turbo-GS~\cite{lu2024turbogs}. We run all approaches on our system except Turbo-GS~\cite{lu2024turbogs}, as the implementation was not available on the day of this submission. We propose three variants of our model, $GSta$, $GSta+TrickGS$ and $GSta+TrickGS+small$, where the first is GSta applied on vanilla 3DGS, second is GSta combined with Trick-GS, and the third is essentially the second but with SH masking for even lower storage. The results are shown in Tables \ref{tab:qual1_2} and \ref{tab:qual_db2_v2}.
 
 \begin{figure*}[t]
\centering
\captionsetup[subfigure]{labelformat=empty}

% \vspace{-0.4em}
\subfloat[GT]{\includegraphics[width=0.325\linewidth]{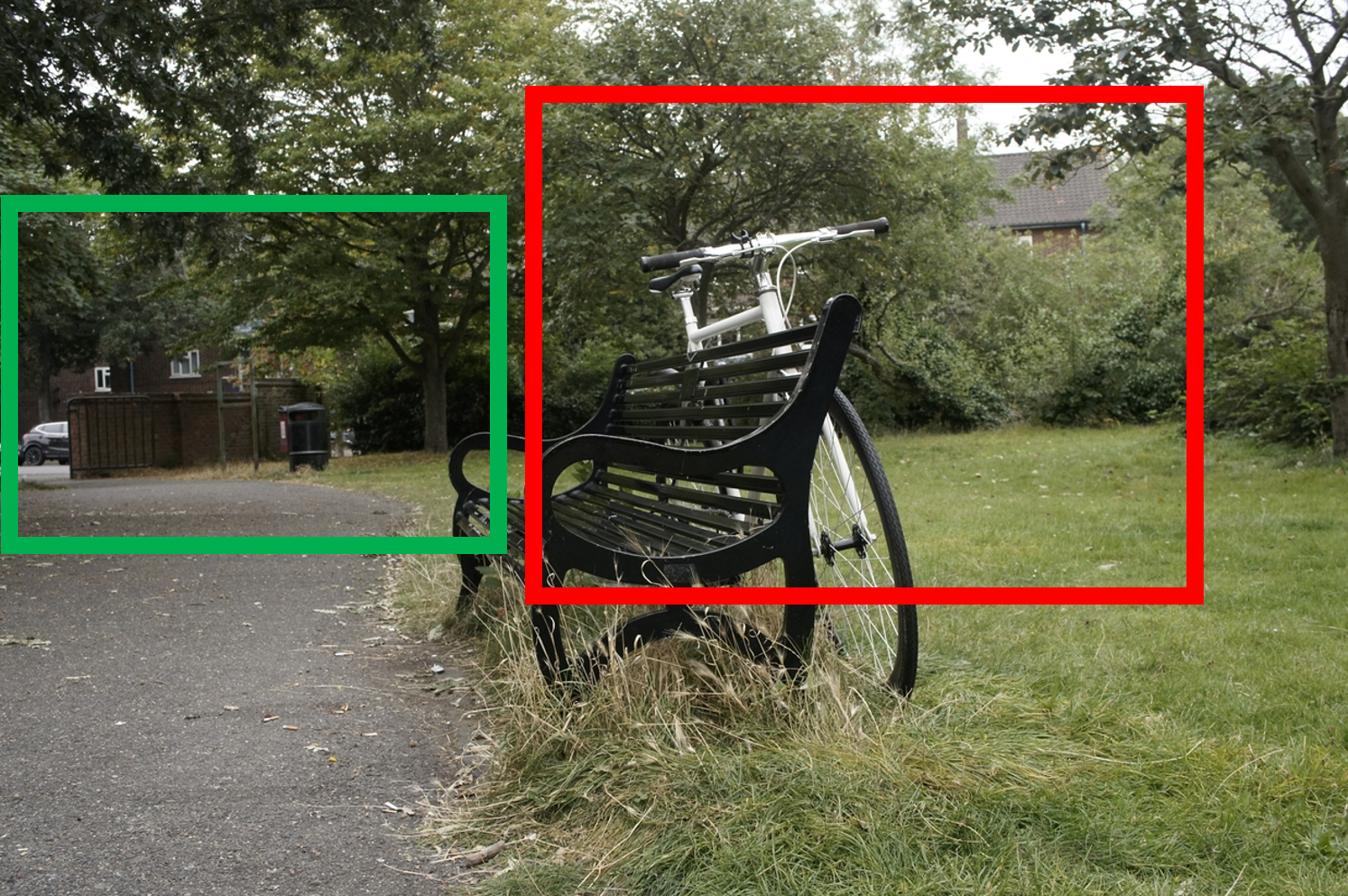}} 
\hspace{0.05em}
\subfloat[\textcolor{green}{Predictions}]{\includegraphics[width=0.325\linewidth]{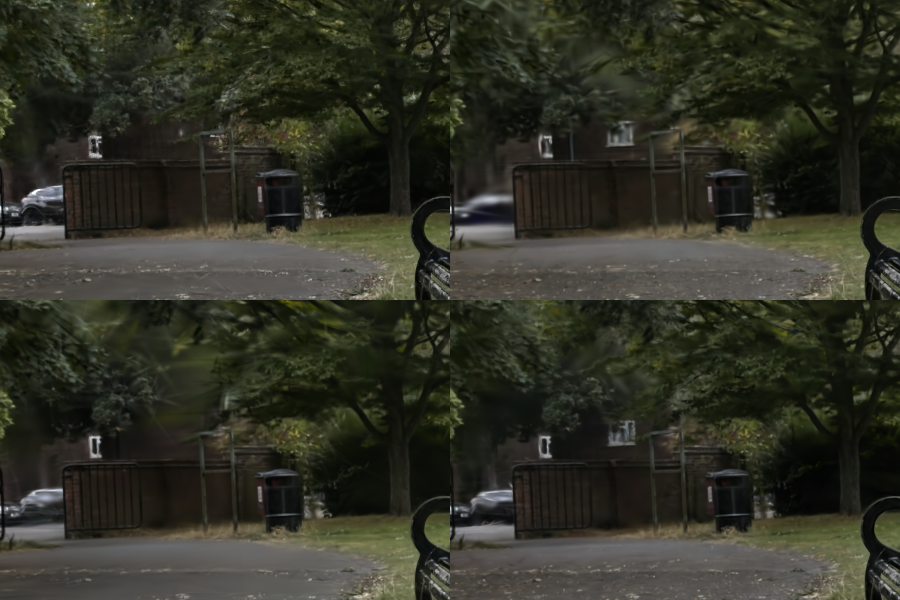}}
\hspace{0.05em}
\subfloat[\textcolor{red}{Predictions}]{\includegraphics[width=0.325\linewidth]{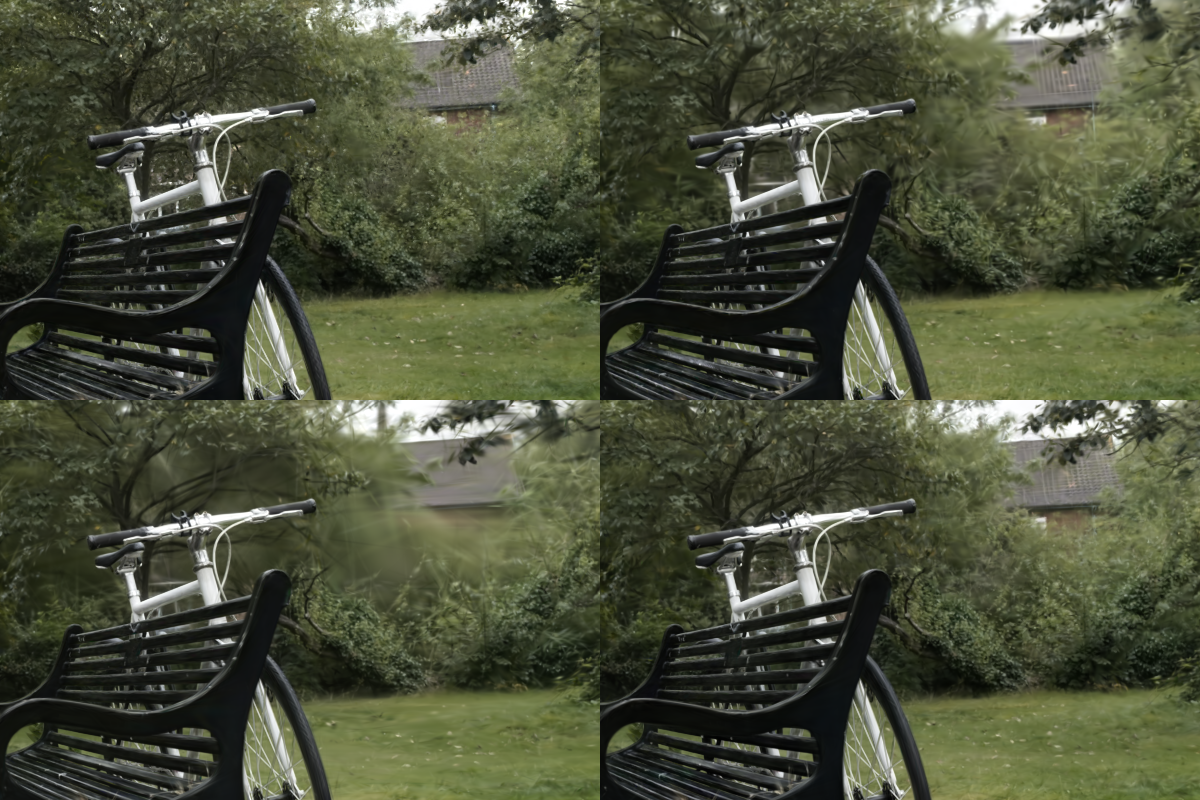}}
\vspace{-1.5mm}
\caption{Qualitative comparison of the methods (top left to bottom right: 3DGS \cite{kerbl3Dgaussians}, Mini-Splatting-v2 \cite{fang2024mini2}, Taming-GS \cite{mallick2024taming3dgs} and ours). Our method can recover more consistent background while keeping low training time and great storage compression rates. We show zoomed prediction images except GT.}
\vspace{-1.5em}
\label{fig:visual_comparison}
\end{figure*}

\noindent \textbf{Results.} $GSta$ achieves competitive accuracy, while reducing the storage requirements and training times $\sim$ 3 fold and improving peak number of Gaussians by 50\% compared to vanilla GS. The results show that $GSta$ achieves faster training than $Trick-GS$ and has half the peak number of Gaussians of $Mini-Splatting-v2$. It has better accuracy (e.g. LPIPS) than $Taming-3DGS$, but exhibits slightly worse storage and peak memory performance. These results show that $GSta$ does improve the current state-of-the-art, and is preferable over others in different aspects; e.g. lower training memory than $Mini-Splatting-v2$, faster than $Trick-GS$ and more accurate than $Taming-3DGS$.

\noindent \textbf{Combining with Trick-GS.} 
$GSta+TrickGS$ presents further gains over $GSta$, where storage, training time and peak memory requirements are all reduced, with little to no drop in accuracy.  Compared to vanilla GS on MipNeRF-360, $GSta+TrickGS$ reduces the storage by $\sim$ 5, training time by $\sim$ 5 and improves peak memory by up to a relative 60\%. Similar to before, $GSta+TrickGS$ now converges even faster than $Trick-GS$, has less than half the peak memory of $Mini-Splatting-v2$ and now trains faster than $Taming-3DGS$. These trends also hold for the other two datasets, where the results show that $GSta$ not only is accurate itself, but also can also leverage orthogonal efficiency improvement methods, such as $Trick-GS$, to achieve even further gains.

$GSta+TrickGS+small$ presents our smallest model, to see how far we can go in achieving efficient GS. At the cost of some drop in accuracy, it introduces some drastic compression rates; it achieves 16$\times$ reduction in storage, 5$\times$ reduction in training time and 2$\times$ reduction in peak memory compared to vanilla 3DGS on MipNerf-360, and similar rates of improvements on other datasets. $GSta+TrickGS+small$ has much less storage size than $Taming-3DGS$ and $Mini-Splatting-v2$, better training time than $Taming-3DGS$ and significantly lower peak memory consumption than $Mini-Splatting-v2$.  In overall, the results show that GSta advances the efficiency-fidelity frontier further and takes a decisive step towards edge-friendly GS systems. 

\noindent \textbf{Qualitative Results.} Images in Figure \ref{fig:visual_comparison} show that our method is highly accurate. $Taming-3DGS$, and $MiniSplatting-v2$ to a certain extent, fail to capture high-frequency details (e.g. tree leaves in both images), whereas our method successfully captures complex details and has no visible difference in quality compared to vanilla 3DGS. 
\vspace{-2mm}
\subsection{Ablation Study}
\vspace{-1mm}
\label{sec:ablation}
We now analyze the impact of GSta and its components. We do so by progressively removing components from GSta, and see their individual effect. The results of our analysis are shown in Table \ref{tab:ablation}. 

\noindent \textbf{Freezing strategies.} We first analyse whether our freezing strategy does improve on two naive approaches; randomly freezing 25\% (\textit{random}) of Gaussians or freezing ones with lowest 25\% gradients (\textit{mingrad}). The results (rows 3, 8 and 9) show that our freezing strategy (GSta) has the best training time. Furthermore, \textit{random} introduces drops in accuracy, as it tends to end training with fewer Gaussians than others because it randomly freezes Gaussians with high gradient magnitudes. Row 3 shows that not using any freezing has the worst training time among others, showing the value of our Gaussian freezing approach. Finally, we evaluate the GSta model without the freeze module trains even slower than using a freeze module with random selections.

\noindent \textbf{LR scheduler and dynamic thresholding.} The addition of LR scheduler (rows 5) shortens the training time by 2 minutes but has no tangible effect on other metrics. The introduction of dynamic thresholding (row 7) in our freezing strategy and densification strategy shows improvements in training time as well, showing its practical usefulness.

\noindent \textbf{Gradient information.} We now analyse the need both $rgb$ and $xyz$ gradients in our freezing approach. Rows 6 shows that only using $xyz$ both harms training time, storage and peak memory. This shows the practical value of \textit{rgb} gradients in being a proxy for a Gaussian's usefulness.   

\noindent \textbf{Trick-GS.} $Trick-GS$ techniques synergizes very well with our GSta (see rows 2 and 3), where huge gains in storage and peak memory are observed, as well as slight gains in training time, with virtually no drop in accuracy metrics.

\begin{table}[!t]
\centering
% \vspace{-0.4em}
\resizebox{1.\linewidth}{!}{
\begin{tabular}{l|cccT{0.07\textwidth}T{0.05\textwidth}T{0.08\textwidth}T{0.06\textwidth}}
\hline
\rowcolor{lightgray} Dataset      & \multicolumn{7}{c}{MipNeRF360 - bicycle}\\ %\cmidrule(lr){2-9}
\hline
Method & PSNR ↑ & SSIM ↑ & LPIPS ↓ & Storage ↓ (MB) & Time ↓ (min) & Max-\#GS ↓ $\times1000$ & \#GS ↓ $\times1000$ \\
% \midrule
\hline

\rowcolor{lightgray}GSta+TrickGS        & 25.23 & 0.758 & 0.248 & 223 & 5.0 & 2512 & 1982  \tabularnewline
GSta+Trick-GS+small  & 25.20 & 0.756 & 0.248 & 82 & 5.6 & 2507 & 1981 \tabularnewline
\rowcolor{lightgray}GSta                & 25.12 & 0.759 & 0.235 & 457 & 6.4 & 4075 & 4060 \tabularnewline
GSta-freezing       & 25.16 & 0.764 & 0.212 & 427 & 15.7 & 3974 & 3795 \tabularnewline
\rowcolor{lightgray}GSta-lrscheduler    & 25.17 & 0.762 & 0.229 & 454 & 11.3 & 4041 & 4036 \tabularnewline
GSta-rgb    & 25.18 & 0.755 & 0.239 & 509 & 8.9 & 4542 & 4520\tabularnewline
\rowcolor{lightgray}GSta-dynamic        & 25.15 & 0.758 & 0.239 & 443 & 7.8 & 3933 & 3933 \tabularnewline
\hline\hline

GSta+mingrad      & 25.25 & 0.765 & 0.223 & 434 & 9.0 &   3981 & 3855 \tabularnewline
\rowcolor{lightgray}GSta+random       & 25.02 & 0.748 & 0.241 & 353 & 13.1 & 3323 & 3140 \tabularnewline

\hline\hline
3DGS    & 25.21 & 0.767 & 0.205 & 713 & 37.6 & 6331 & 6331 
\\\hline
\end{tabular}}
% \vspace{-0.3em}
\caption{\small Ablation study on different components of our model. ``-" indicates the removal of a component (e.g. -freezing means no freezing) and similarly ``+`` indicates additional modules (e.g. +small means with SH masking). \textit{mingrad} and \textit{random} are alternative ways of freezing (see Section \ref{sec:ablation}).}

% We report the results by enabling or disabling a feature from our GSta model.}
\vspace{-1em}

\label{tab:ablation}
\vspace{-1.5em}
\end{table}

\vspace{-1mm}
\section{Conclusion}
\vspace{-1mm}
On device learning of GS is costly and it requires efforts on multiple evaluation metrics, such as  accuracy, training time, storage requirement. In this work, we propose GSta, a generic training scheme for 3DGS that can be used as a plug-and-play module for any GS-based method. Consisting of a novel Gaussian freezing scheme, early stopping mechanism and improved hyper-parameter selections, GSta advances the frontier of efficient GS methods on multiple fronts, such as training memory, time and storage requirements, while preserving competitive accuracy.
% \clearpage
\bibliographystyle{IEEEtran}
\bibliography{main}

\end{document}